%% file: main.tex
\documentclass[pdflatex,sn-mathphys-num]{sn-jnl}

\usepackage{graphicx}%
\usepackage{multirow}%
\usepackage{amsmath,amssymb,amsfonts}%
\usepackage{amsthm}%
\usepackage[title]{appendix}%
\usepackage{xcolor}%
\usepackage{textcomp}%
\usepackage{manyfoot}%
\usepackage{booktabs}%
\usepackage{algorithm}%
\usepackage{algorithmicx}%
\usepackage{algpseudocode}%
\usepackage{listings}%
\usepackage{array}
\usepackage[caption=false,font=normalsize,labelfont=sf,textfont=sf]{subfig}
\usepackage{textcomp}
\usepackage{stfloats}
\usepackage{url}
\usepackage{verbatim}
\usepackage{hyperref} 
\usepackage{enumitem}
\usepackage{color}         
\usepackage{tabularray}
\usepackage{bbm}

\theoremstyle{thmstyleone}%
\theoremstyle{thmstyletwo}%

\theoremstyle{thmstylethree}%

\begin{document}

\title{Hyperpruning: Efficient Search through Pruned Variants of Recurrent Neural Networks Leveraging Lyapunov Spectrum}


\author[1]{\fnm{Caleb} \sur{Zheng}}\email{zheng94@uw.edu}

\author*[1,2]{\fnm{Eli} \sur{Shlizerman}}\email{shlizeeuw.edu}

\affil[1]{\orgdiv{Department of Electrical and Computer Engineering}, \orgname{University of
Washington}, \orgaddress{\city{Seattle}, \postcode{98195}, \state{WA}, \country{USA}}}

\affil[2]{\orgdiv{Department of Applied Mathematics}, \orgname{University of
Washington}, \orgaddress{\city{Seattle}, \postcode{98195}, \state{WA}, \country{USA}}}


\abstract{A variety of pruning methods have been introduced for over-parameterized Recurrent Neural Networks to improve efficiency in terms of power consumption and storage utilization. These advances motivate a new paradigm, termed `hyperpruning', which seeks to identify the most suitable pruning strategy for a given network architecture and application. Unlike conventional hyperparameter search, where the optimal configuration's accuracy remains uncertain, in the context of network pruning, the accuracy of the dense model sets the target for the accuracy of the pruned one. The goal, therefore, is to discover pruned variants that match or even surpass this established accuracy. However, exhaustive search over pruning configurations is computationally expensive and lacks early performance guarantees. To address this challenge, we propose a novel Lyapunov Spectrum (LS)-based distance metric that enables early comparison between pruned and dense networks, allowing accurate prediction of post-training performance. By integrating this LS-based distance with standard hyperparameter optimization algorithms, we introduce an efficient hyperpruning framework, termed \textit{LS}-based \textit{H}yperpruning (\textbf{\textit{LSH}}). LSH reduces search time by an order of magnitude compared to conventional approaches relying on full training. Experiments on stacked LSTM and RHN architectures using the Penn Treebank dataset, and on AWD-LSTM-MoS using WikiText-2, demonstrate that under fixed training budgets and target pruning ratios, LSH consistently identifies superior pruned models. Remarkably, these pruned variants not only outperform those selected by loss-based baseline but also exceed the performance of their dense counterpart.}




\maketitle

\section{Introduction}
Over the past decade, Recurrent Neural Networks (RNNs) have achieved significant performance improvement across a wide range of sequence modeling tasks, including action recognition~\cite{su2020predict} and video summarization~\cite{zhao2018hsa} and voice conversion~\cite{huang2021pretraining}. In particular, RNN variants, such as LSTM~\cite{hochreiter1997long, malhotra2015long}, and RHN~\cite{zilly2017recurrent}, are plausible neural architecture for various natural language processing tasks, including machine translation~\cite{wu2016google} and language modeling~\cite{irie2019language}. However, the computational demands of RNNs scale linearly with input sequence length and quadratically with network size. This leads to slow training and inference, which poses challenges for deploying RNNs on resource-constrained platforms, such as mobile devices.

Various techniques have emerged to tackle the increasing computational cost, including network pruning~\cite{han2015learning,narang2017exploring, zhu2017prune, gale2019state},  quantization~\cite{hernandez2020hybrid, han2015deep}, weight sharing~\cite{ullrich2017soft}, and knowledge distillation~\cite{gou2021knowledge}. Among these techniques, network pruning offers a particularly promising avenue, aiming to achieve a pruned network, i.e., a network that is more sparse and thus requires fewer computational resources and less storage. A widely used approach is dense-to-sparse training, which gradually removes weights from a dense network during the training process. While this inference time of pruned networks eventually decrease, the training time often remains similar or, in some cases, becomes even longer than the training of a dense network. This prolonged training time, often over a span of multiple days or weeks, presents substantial challenges in efficiently achieving pruned networks.

Recent efforts have introduced the Dynamic Sparse Training (DST) to cater to the growing demand for optimizing computational costs in achieving pruned variants. In contrast to dense-to-sparse approaches, DST adopts a sparse-to-sparse pruning method, being initiated with a sparse network and maintaining a consistent number of non-zero parameters throughout training. This approach seeks to enhance not only the inference time but also the duration of training. DST involves three pivotal steps in pruning: weight removal, weight growth, and weight redistribution. Each step is controlled by salience criterial (magnitude- or gradient-based rules) to determine their execution. Notably, no universal set of controls is applicable across all tasks and architectures, as each scenario presents its unique requirements. This unique mapping between a scenario and a control in DST prevents generalizing a standard rule over all scenarios. Therefore, for a particular scenario, the controls that characterize the pruning method become additional key hyperparameters to be set to execute pruning optimally. We term this type of hyperparameter search as \textbf{'Hyperpruning'}.

\begin{figure*}[ht]
    \centering
    \includegraphics[width=\linewidth]{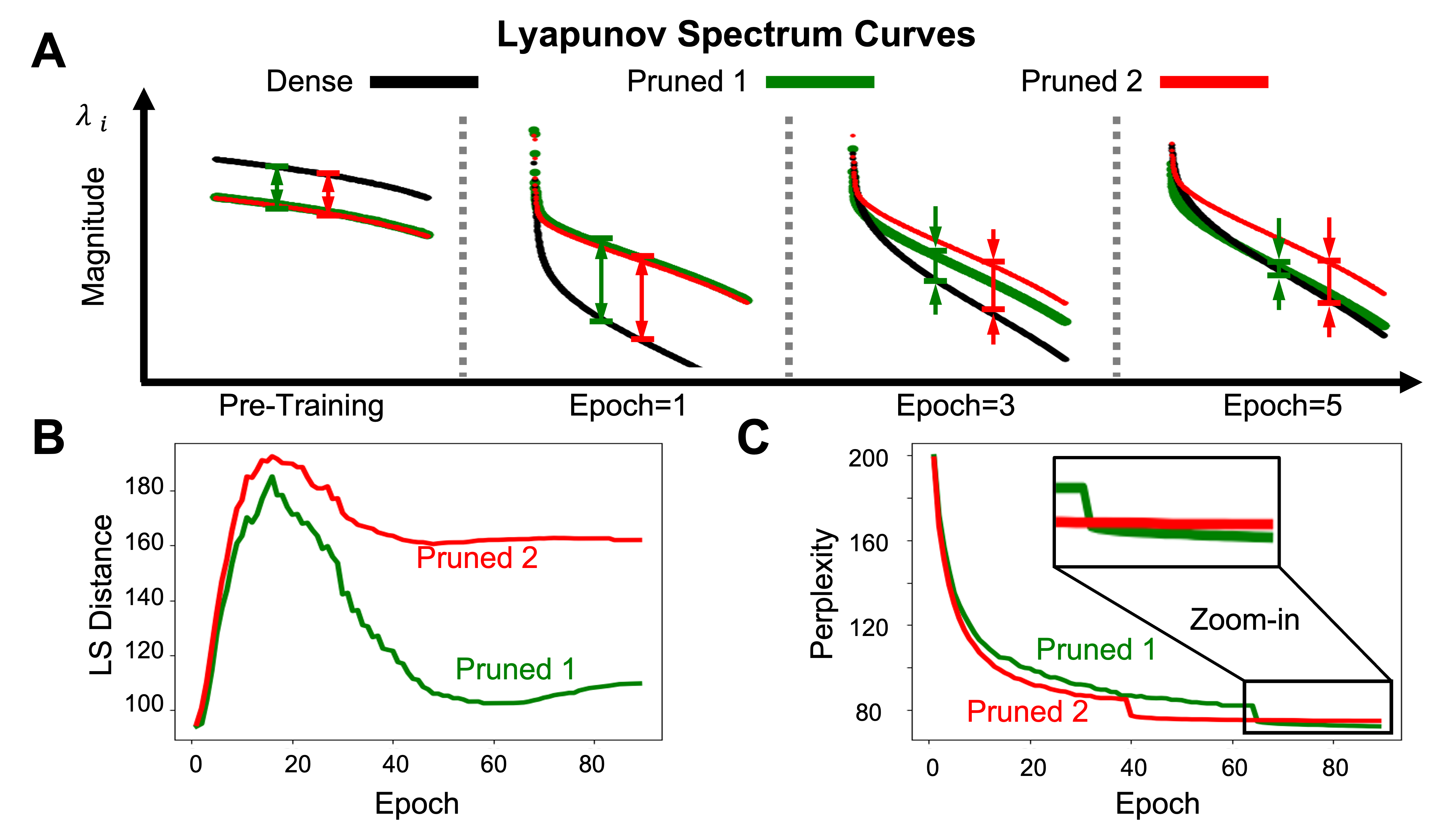}
    \caption{(A) Lyapunov Spectrum (LS) curves of the dense network (black) and two pruned variants—Pruned 1 (green) and Pruned 2 (red)—evaluated at pre-training and at epochs 1, 3, and 5 (left to right). The dense network serves as a reference trajectory in LS space. Arrows indicate the deviation magnitude from the dense model. (B) Temporal evolution of the $L_2$ distance between each pruned variant and the dense network in LS space throughout training.  (C) Perplexity curves of the two pruned variants over training epochs.}
    \label{fig:LS_Explained}
\end{figure*}

Hyperpruning involves selecting the pruning methodology with its controls and other hyperparameters related to training for a specific scenario. This process necessitates searching for a configuration that includes methodological and non-methodological hyperparameters. Methodological hyperparameters define the specifics of the pruning method, while non-methodological hyperparameters are independent of the pruning methods and are applicable across them. A distinguishing feature of hyperpruning, setting it apart from many other hyperparameter search problems, is the availability of an estimated target accuracy. This estimation is possible due to the knowledge of the accuracy of the dense counterpart, which serves as a loose upper bound to guide the hyperpruning process. However, even with this knowledge, searching through the configuration variants of different pruning methods and their associated hyperparameters remains time-consuming, as each variant requires extensive training. Furthermore, there is no guarantee that a more accurate variant can be achieved after multiple iterations of unsuccessful configuration variants. 

Hyperparameter optimization algorithms expedite the search process by leveraging a distance metric, enabling efficient \textit{evaluation} of configurations or the \textit{generation} of reliable configurations. This distance metric aims to provide an early estimation of the accuracy achievable by a particular configuration without performing full training. Typically, the loss curve from the early stages of training is used as a distance metric. However, it falls short in multiple scenarios, as it does not guarantee that a model with low loss in the initial training phase will reach the desired accuracy in the end, as illustrated in Fig.~\ref{fig:LS_Explained}-C. Therefore, the development of a predictive distance metric is key for hyperpruning.

In the context of sequence networks, including RNNs and their variants, prior research has demonstrated that these networks can be treated as dynamical systems, and their dynamic flow can be analyzed using the Lyapunov Spectrum (LS). The LS measures the contraction/expansion ratio of the network states over time and has been proposed as an indicator of network trainability and achievable accuracy~\cite{vogt2022lyapunov}. Recent work has suggested that an AutoEncoder learned from the LS of slightly trained RNN samples with random hyperparameter configurations can establish a low-dimensional embedding space where RNN variants are organized according to their post-training accuracy~\cite{vogt2022lyapunov}. This organizational structure appears to persist in the pruning setup when creating the embedding directly from LS, as evidenced in ~\ref{fig:LS_Explained}-A. The example compares two pruned variants (green and red) with the dense network (black). After full training, Pruned 1 variant (green) eventually achieves more optimal/lower perplexity than Pruned 2 variant (red). However, its perplexity is higher for more than 60 epochs, see Fig.~\ref{fig:LS_Explained}-C. While early perplexity is not indicative of the estimated accuracy, LS-based distance, on the other hand, appears to be consistent and indicates that the Pruned 1 variant is closer to the dense network for all epochs, as shown in Fig.~\ref{fig:LS_Explained}-B.

The consistency of the distance between the dense and pruned variants throughout training motivates the introduction of a novel distance metric based on LS and its embedding (LS Space) to guide hyperpruning. Alongside the LS distance, we propose the \textit{LS}-based \textit{H}yperpruning (\textbf{\textit{LSH}}) algorithm, which utilizes the LS-based distance as an early estimation criterion. In summary, our contributions are as follows:

\begin{enumerate}[leftmargin=*]
\item We propose a novel distance metric for hyperpruning based on the Lyapunov Spectrum (LS), which is capable of estimating the similarity between a pruned network and its dense counterpart.
\item We introduce LSH, a hyperpruning algorithm that utilizes the LS-based distance as an early estimation criterion and allows sifting through and replenishing pruned variants with an order of magnitude smaller number of training epochs than loss-based full training search.
\item Our extensive experiments on RNN language model benchmarks, including stacked LSTM and RHN models trained on the Penn Treebank dataset and AWD-LSTM-MoS trained on the WikiText-2 dataset, demonstrate that LSH outperforms conventional loss-based hyperparameter optimization search methods, state-of-the-art pruning methods, and even surpasses the performance of the dense networks.
\end{enumerate}

\section{Related Work}

\subsection*{Network Pruning}
Network pruning is a fundamental approach for reducing model complexity and computational cost in machine learning~\cite{hoefler2021sparsity, wang2021recent}. Early applications of pruning focused on enhancing inference speed by pruning pre-trained networks and then retraining them to recover accuracy losses incurred due to pruning~\cite{janowsky1989pruning, mozer1989using, mozer1988skeletonization, lecun1989optimal, hassibi1992second}. Such pipelines are referred to as \textit{dense-to-sparse} and are applied across various domains. For example, magnitude-based pruning was employed to eliminate weights of less significance, achieving near-comparable accuracy to dense networks in multiple instances~\cite{han2015learning, han2015deep, li2016pruning, guo2016dynamic}. Subsequently, additional criteria were developed to assess the importance of weights~\cite{dubey2018coreset, hu2016network, tan2020dropnet, molchanov2016pruning, molchanov2019importance, lin2020hrank,chin2020towards}. Several methods eliminate the need for a pre-trained network by training a network from scratch while concurrently engaging in direct or indirect model pruning. Direct pruning methods include Gradual Magnitude Pruning (GMP)~\cite{narang2017exploring,  zhu2017prune, gale2019state}, as well as techniques such as weights and pruning mask joint optimization~\cite{liu2020dynamic, kusupati2020soft, lin2020dynamic}. Indirect pruning approaches include $L_0$~\cite{louizos2017learning}, $L_1$ regularization~\cite{wen2017learning}, and Variational Dropout~\cite{molchanov2017variational}. However, these methods often fall short in large-scale learning tasks compared to magnitude-based pruning~\cite{gale2019state}. Moreover, they can incur a training time penalty that sometimes exceeds the time required for training a dense network.

To enhance training efficiency, \textit{Sparse-to-sparse} pruning methods gained traction. These methods begin with a sparse initialization and maintain a constant number of non-zero parameters throughout training. The Lottery Ticket Hypothesis shows that such a non-trivial sparse network, the so-called "lottery ticket", exists and can be trained to match the performance of a dense network. The "lottery ticket" might even surpass dense networks due to extra regularization of pruning~\cite{frankle2018lottery}. The identification of "lottery tickets" at an early stage of training was explored in \cite{lee2018snip, lee2019signal, wang2020picking, tanaka2020pruning, you2019drawing}. These methods typically employ static masks, which fix the location of non-zero parameters over the entire training and do not match up with the outcomes of dense-to-sparse methods~\cite{wang2020picking} and do not perform well in extremely sparse scenarios~\cite{lee2019signal}.

Dynamic Sparse Training (DST) emerged as a solution to replace static masks with dynamic ones, enhancing the flexibility and the outcome accuracy of the pruned network~\cite{bellec2017deep, dai2019nest}. Specifically, DST allows the alteration of non-zeros parameter positions through a prune-regrow process, which maintains a constant total number of zero and non-zero weights. Deep-R first introduced this idea by rewiring the network from a posterior standpoint~\cite{bellec2017deep}. SET simplified this process using magnitude-based pruning and random weight growing~\cite{mocanu2018scalable}. DSR improved upon it by introducing non-uniform sparsity across different layers~\cite{mostafa2019parameter}. While promising, sparse-to-sparse methods still lagged behind dense-to-sparse methods. SNFS proposed to use momentum for weight growth to match or even outperform dense-to-sparse methods~\cite{dettmers2019sparse}, albeit at a computational cost~\cite{dettmers2019sparse}. To address this inefficiency, RigL introduced ``lazy gradient" calculation into magnitude-based pruning~\cite{evci2020rigging}, and Top-KAST~ further optimized it by eliminating the need to calculate dense gradients\cite{jayakumar2020top}. A DST method called Selfish-RNN has been explicitly proposed for RNN and showed a significant performance improvement via a non-uniform weight redistribution across gates and SNT-ASGD~\cite{liu2021selfish}. Selfish-RNN addresses parameter allocation among different layer types, which was also observed as key in other works~\cite{kusupati2020soft, frankle2020pruning}. Additional, comprehensive surveys on network pruning have been made available in~\cite{hoefler2021sparsity, wang2021recent} and references therein.

\subsection*{Hyperparameter Search}
HyperParameter Optimization (HPO) algorithms aim to identify the optimal configuration of model hyperparameters to maximize performance. HPO algorithms can generally be divided into two categories: configuration \textit{proposal} and configuration \textit{evaluation}.

Random search is a widely used method for proposing configurations and is typically more efficient than grid search~\cite{bergstra2012random}. Advanced HPO algorithms enhance configuration proposal efficiency by maintaining an archive of proposed configurations and their corresponding performance. These algorithms dynamically choose between exploitation (local search, exploiting the current best configuration) and exploration (random search, exploring new configurations). For example, Bayesian Optimization, such as Tree Parzen Estimators (TPE) and Adaptive TPE (ATPE), utilizes the archive to fit a surrogate model and adapts the proposition of configurations~\cite{snoek2012practical, hutter2011sequential, bergstra2011algorithms}. Given this extra information from the archive, Bayesian Optimization has demonstrated improved performance compared to plain random search~\cite{thornton2013auto, eggensperger2013towards, snoek2015scalable}.

Configuration evaluation involves an efficiency-accuracy trade-off, with more computational resources leading to more accurate evaluations. Multi-fidelity search aims to rank candidates based on certain distances, allowing for the allocation of additional resources to promising candidates and the removal of non-promising ones~\cite{bischl2021hyperparameter}. Multi-fidelity search can be integrated into random search or Bayesian optimization algorithms to explore a more expansive searching space within the same resource budget~\cite{li2017hyperband, falkner2018bohb}. Fast and effective configuration evaluation methods become essential in complex architectures, such as sparse stacked LSTM, RHN, and AWD-LSTM-MoS, with numerous potential hyperparameter configurations. We argue that the Lyapunov Spectrum (LS) holds promise, efficiently assessing candidates early in training.

\subsection*{Lyapunov Spectrum}
Treating sequence networks, such as RNN, as dynamical systems has emerged as a compelling approach in deep learning research for understanding and predicting long-term network behavior~\cite{chang2019antisymmetricrnn, zheng2020r, 10.3389/fams.2022.818799, ribeiro2020beyond}. Lyapunov Spectrum (LS), composed of Lyapunov Exponents, was initially employed to gain insights into autonomous neural networks~\cite{monteforte2010dynamical, engelken2020lyapunov}. Despite RNNs being non-autonomous due to external inputs, LS can be generalized to them based on random dynamical theory as long as input sequences are sampled from a stationary distribution~\cite{arnold1995random}. Another property of LS is its convergence to the same curve if the RNN sequence is long enough, as proven by Oseledets theorem~\cite{saito1979ergodic, ochs1999stability}. This property implies that LS captures intrinsic dynamical characteristics of the network. Prior studies have used the sign of the largest Lyapunov exponent as an indicator of chaotic dynamics~\cite{legenstein2007edge, pennington2018emergence, laurent2016recurrent}. Furthermore, additional LS features, such as zeros, negativity, mean, and variance, have been linked to concepts in dynamical systems, including quasi-periodic orbits, fixed-point attractors, heterogeneity, and rates of contraction~\cite{dawson1994obstructions, abarbanel1991variation, shibata2001ks, brandstater1983low, yamada1988inertial}. While each feature seems to extract properties of the network, the correlation between these features and network accuracy remains unclear~\cite{10.3389/fams.2022.818799}. 

A significant advancement was made in \cite{10.3389/fams.2022.818799}, where an efficient method for computing the full Lyapunov Spectrum was introduced. Building on this, AeLLE demonstrated that LS can be leveraged to predicted the generalization performance of RNNs with randomly sampled hyperparameters~\cite{vogt2024lyapunov}. In particular, an autoencoder trained on LS representations was shown to learn a low-dimensional embedding space where network variants are organized according to their post-training accuracy.

Building on this innovative foundations, we propose an LS-based distance metric to enhance the efficiency of hyperpruning. By treating dense model as a reference and measuring the distance between its LS and that of a pruned variant, we obtain a robust indicator of dynamical similarity. This distance enables the early elimination of suboptimal candidates and generation of promising new candidates. As a result, our LS-guided approach offers a principled and computationally efficient strategy for improving the efficiency and effectiveness of network pruning procedures.

\begin{figure*}[ht]
    \centering
    \includegraphics[width=1.0\linewidth]{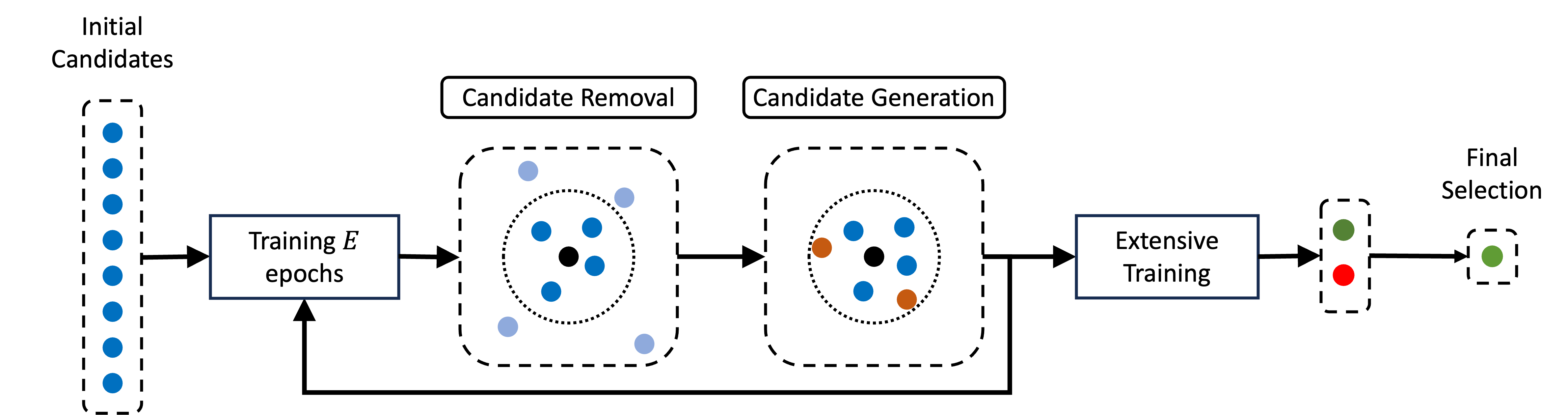}
    \caption{LS-based Hyperpruning: starts with a set of candidates (blue) and a pre-trained dense reference model (black), LSH executes candidate removal and generation. \textbf{Candidate Removal}: Each candidate is trained for $E$ epochs and projected to the LS space, along with the reference. Candidates with further distance (light blue) are removed based on their distance to the reference. \textbf{Candidate Generation}: New candidates (brown) are adaptively generated based on remaining candidates and inserted into the candidate set. After several iterations of removal and generation, the remaining candidates in the set are extensively trained and the best one (green) is picked as the final selection.}
    \label{fig:selection_process}
\end{figure*}

\section{Methods}
In this section, we describe the computation of the Lyapunov Spectrum (LS) and describe the proposed LS-based Hyperpruning (LSH) algorithm. As illustrated in Fig.~\ref{fig:selection_process}, LSH begins with a candidate pool of pruned network configurations and iteratively eliminates non-promising candidates based on their LS distance to a reference dense network. New candidates are then generated using adaptive sampling. After several rounds of candidate removal and generation, the most promising configurations are selected for full training. Notably, the LS-based distance metric in LSH can also be integrated with existing hyperparameter optimization algorithms.

\subsection*{Lyapunov Spectrum Computation}
Lyapunov Exponents (LEs) are used in dynamical systems to quantify the rate of contraction and expansion of infinitesimally close trajectories. Specifically, for two infinitesimally close points separated by an initial distance $\epsilon$, their divergence after $T$ time steps evolves as $\epsilon e^{\lambda T}$, where $\lambda$ is the Lyapunov Exponent. For RNNs, which evolve over hidden states, the LS comprises a set of LEs equal in number to the dimension of the hidden state. Although RNNs are non-autonomous due to their input sequences, LS can still be computed by invoking random dynamical systems theory, assuming inputs are drawn from a stationary distribution.

Following the procedure in \cite{vogt2024lyapunov}, LS is computed by tracking the contraction/expansion of the hidden states of the RNN over time. Specifically, for a batch $\mathcal{X}$ comprising $K$ samples, each with a sequence length of $T$, from the validation set $\mathcal{D}_{val}$. Symbolically, $\mathcal{X}$ is represented as $\{ \textbf{x}^k = \{x^{t, k}\}_{t=1}^T, x^{t,k} \in \mathcal{R}^d, k = 1, 2, \dots, K\}$, where $x^{t, k}$ denotes the $t$-th time step of the $k$-th sample in $\mathcal{X}$, and $d$ is the input dimension. Initialization includes setting the hidden states of the RNN and an orthogonal representation base at time $0$ as vectors of zeros and the identity matrix \textbf{$\mathcal{I}$}, respectively ($h^0$ and $\mathcal{B}^0$). At each time step ($t$), the hidden states ($h^t$) track the network evolution, while the orthogonal representation base ($\mathcal{B}^t$) captures the stretching of each direction in the hidden states up to time step $t$. These quantities are updated as follows $h^{t, k} = f(h^{t-1, k}, x^{t, k})$ and $\mathcal{B}^{t, k}, R^{t, k} = qr(J^{t, k} \cdot \mathcal{B}^{t-1, k})$, where $f$ represents the network function, $J$ is the Jacobian matrix between hidden states of adjacent time steps, $qr$ denotes the QR decomposition, and $(\cdot)$ denotes matrix multiplication.

The derivation of the Jacobian matrix for all architectures used in this paper is shown in Appendix.~\ref{sm:jacobian}. The stretching ratio at time step $t$ is incorporated into the $i-$th column of $\mathcal{B}^t$ and measured by $r^{t}_i$, representing the $i$-th diagonal element of $\mathcal{R}^t$. The $i$-th $LE$ ($\lambda_i$) is then obtained by averaging over the entire sequence $T$ and $K$ samples, as defined by Equation~\ref{eq:LE_cal}: 
\begin{align}
    \lambda_i & = \frac{1}{K} \frac{1}{T}\sum^K_{k=1} \sum^T_{t=1} log(r_{i}^{t, k}).
    \label{eq:LE_cal}
\end{align}
We denote the LS as $\Lambda$, such that $\Lambda \equiv \{\lambda_i\}_{i=1}^N$, where $N$ is the dimension of the hidden states $h^t$.

\subsection*{LS-based Hyperpruning (LSH)}
The LSH algorithm aims to efficiently identify the optimal pruned RNN configuration. It starts with a candidate pool $\mathcal{P}$, where each candidate $c^i \in \mathcal{P}$ represents a pruned network configuration. The LSH procedure is guided by a pre-trained dense reference network, denoted as $\hat{c}$, which serves as a target for the pruned networks to reach and, in some cases, even surpass. Candidates are selected based on LS-based removal and adaptive generation, executed iteratively, as shown in Fig.~\ref{fig:selection_process}.

During the LS-based removal step, each candidate is trained for $E$ epochs. After each epoch, the LS of each candidate is computed using Eq.~\ref{eq:LE_cal}, resulting in $\Lambda^i_j$, where $i$ and $j$ represent the $i$-th candidate and the $j$-th epoch, respectively. The LS history of the trained candidate $c^i$ is denoted as $\Lambda^i \equiv \{\Lambda^i_j\}_{j=0}$, while the LS of the dense network is denoted as $\hat{\Lambda}$. $\Lambda^i$ and $\hat{\Lambda}$ are projected onto an embedding space, and the distance between $\Lambda^i_{-1}$ and $\hat{\Lambda}$ is used to quantify the closeness between the pruned candidate and the dense network, where $_{-1}$ denotes the current epoch.

Principal Component Analysis (PCA) is chosen to construct a low-dimensional (2D) embedding space, termed LS Space, into which LS is projected. The $L_2$ distance in this space is used to assess the closeness of networks and rank candidates in $\mathcal{P}$. However, there are no special constraints on this choice, and the combination of other embedding spaces and distances is discussed in the Appendix.~\ref{sm:embed}. In LSH method, candidates with further distances to the dense network reference are removed, with $\frac{n}{2}$ candidates selected from $n$ candidates in $\mathcal{P}$. It is important to note that LS-based removal efficiently identifies the best configuration within $\mathcal{P}$, but there is no guarantee that the best configuration exists in the initial pool. Therefore, in addition to removal, LSH adaptively generates new candidates based on the remaining candidates in $\mathcal{P}$, gradually refining the distribution of candidates to improve the chances of finding the best configuration. For adaptive generation, existing Bayesian Optimization algorithms, such as TPE and ATPE, are adapted to generate new candidates, leveraging the knowledge of remaining candidates. Specifically, $\frac{n}{4}$ new candidates are generated and added to the candidate set $\mathcal{P}$. This imbalance in the number of removal ($\frac{n}{2}$) and generation ($\frac{n}{4}$), results in an asymptotic reduction in the number of candidates in $\mathcal{P}$, ultimately keeping the more promising candidates. The hyperparameter $E$ determines when to execute candidate removal and generation, thus influences the time budget allocated for the candidate selection process. Following $m$ epochs of the candidate selection process, the remaining candidates in $\mathcal{P}$ undergo extensive training until their losses/accuracy converge. Subsequently, the best-performing candidate is selected as the final pruned network configuration.

\section{Experiments}
\label{sec:result}
In this section, we validate the performance and the generality of the proposed LS-based Hyperpruning (LSH) method across various RNN architectures and datasets. We evaluate stacked LSTM and RHN models on the Penn TreeBank Dataset~\cite{marcinkiewicz1994building}, and AWD-LSTM-MoS on the WikiText-2 Dataset~\cite{merity2016pointer} for language modeling tasks. To ensure comparability, we adopt the same experimental setup as in prior work~\cite{liu2021selfish} for consistency. We also examine the robustness of LSH by conducting experiments on stacked LSTM under different pruning ratios. Furthermore, we measure the time efficiency of LSH in comparison to traditional loss-based full training search, reporting the time required to achieve a given target accuracy. Finally, we evaluate LSH under five different HPO algorithms to demonstrate its effectiveness as an early-estimation criterion compared to loss-based distance. All experiments are conducted on an NVIDIA RTX 2080 Ti. Additional architectural and hyperparameter details are provided in Appendix.~\ref{sm:hyperparameter}.

\begin{table}[t]
\centering
\caption{The testing perplexity for language modeling task on Penn TreeBank Dataset with stacked LSTM and RHN, and on WikiText-2 Dataset with AWD-LSTM-MoS. DSR, RigL, SNFS, SET, ISS, GMP, Dense, and Selfish-RNN data are from~\cite{liu2021selfish}.}
\begin{tabular}{l|cc|cc|cc}
\toprule
\textbf{Datasets} & \multicolumn{4}{c|}{\textbf{PTB}}                                                     & \multicolumn{2}{c}{\textbf{Wiki-2}}            \\
\midrule
Models                 & \multicolumn{2}{c|}{Stacked-LSTM}     & \multicolumn{2}{c|}{RHN}              & \multicolumn{2}{c}{AWD-LSTM-MoS}      \\

                       & \# Param                & Perplexity & \# Param                & Perplexity & \# Param                & Perplexity  \\
\midrule
Dense                  & 66.0M                   & 78.6       & 23.5M & 65.4 & - & -\\
Dense (NT-ASGD)        & 66.0M & 72.4       & 23.5M& 61.8 & 35.0M & 63.3 \\ 
\midrule
 & \multicolumn{2}{c|}{Sparsity = 0.67} & \multicolumn{2}{c|}{Sparsity = 0.53} &  \multicolumn{2}{c}{Sparsity = 0.55} \\
\midrule
DSR (Adam)             & \multirow{9}{*}{21.8 M} & 88.2       & \multirow{9}{*}{11.1 M} & 63.2       & \multirow{9}{*}{15.6 M} & 67.0        \\
RigL (Adam)            &                         & 85.6       &                         & 64.4       &                         & 68.5        \\
SNFS (Adam)            &                         & 86.3       &                         & 71.1       &                         & 76.2        \\
SET (Adam)             &                         & 85.5       &                         & 61.1       &                         & 69.6        \\
ISS                    &                         & 78.7       &                         & 65.4       &                         & -           \\
RigL (SNT)        &                         & 75.9       &                         & 62.5       &                         & 65.2        \\
GMP (SNT)         &                         & 74.8       &                         & 64.0       &                         & -           \\
Selfish-RNN (SNT) &                         & 71.7       &                         & 60.4       &                         & 63.0        \\
\textbf{LSH (Ours)} & & \textbf{69.9} & & \textbf{59.0} & & \textbf{62.2} \\
\bottomrule
\end{tabular}
\label{table:result}
\end{table}

\subsection*{Effectiveness}
To assess the effectiveness of LSH, we compared its results with those of dense networks and other state-of-the-art pruning methods. We applied LSH to three benchmarks (stacked LSTM, RHN trained on PTB Dataset, and AWD-LSTM-MoS on WikiText-2 Dataset with pruning ratios of $0.67$, $0.53$, and $0.55$, respectively). 

During hyperpruning, LSH searches over both methodological and non-methodological hyperparameters. We considered sparse initialization, death, and redistribution modes as methodological hyperparameters. Sparse initialization determines the initial layer-wise pruning ratio for each layer. Death and redistribution modes decide how the weights in the network are pruned and redistributed before and after weight growth. We provided $2$ available initialization modes, $4$ death modes, and $3$ redistribution modes, resulting in $24$ variants. In addition, for non-methodological hyperparameters, we select the death rate, representing the ratio of pruned weight to the total non-zeros weight. It is selected from a continuous range $[0.4, 0.9]$.

Our observations reveal that LSH identified non-trivial hyperparameter configurations which varied across different architectures. For example, the selected stacked LSTM candidate has a non-trivial death rate ($0.58$) compared to the empirical value used in the SOTA method Selfish RNN ($0.8$). The sparse initialization modes for the selected stacked LSTM and AWD-LSTM-MoS candidates differed, with stacked LSTM using uniform initialization and AWD-LSTM-MoS employing ER initialization. As a result of this selection process, LSH consistently outperformed dense networks and other state-of-the-art pruning methods, as demonstrated in Table~\ref{table:result}. Specifically, LSH improves the previous best results (Selfish RNN) in all three scenarios: $69.9$ vs. $71.7$ for Stacked LSTM, $59.0$ vs. $60.4$ for RHN, and $62.2$ vs. $63.0$ for AWD-LSTM-MoS. Notably, these improvements were achieved purely through hyperparameter search.
\begin{figure*}[t]
    \centering
\includegraphics[width=1.0\linewidth]{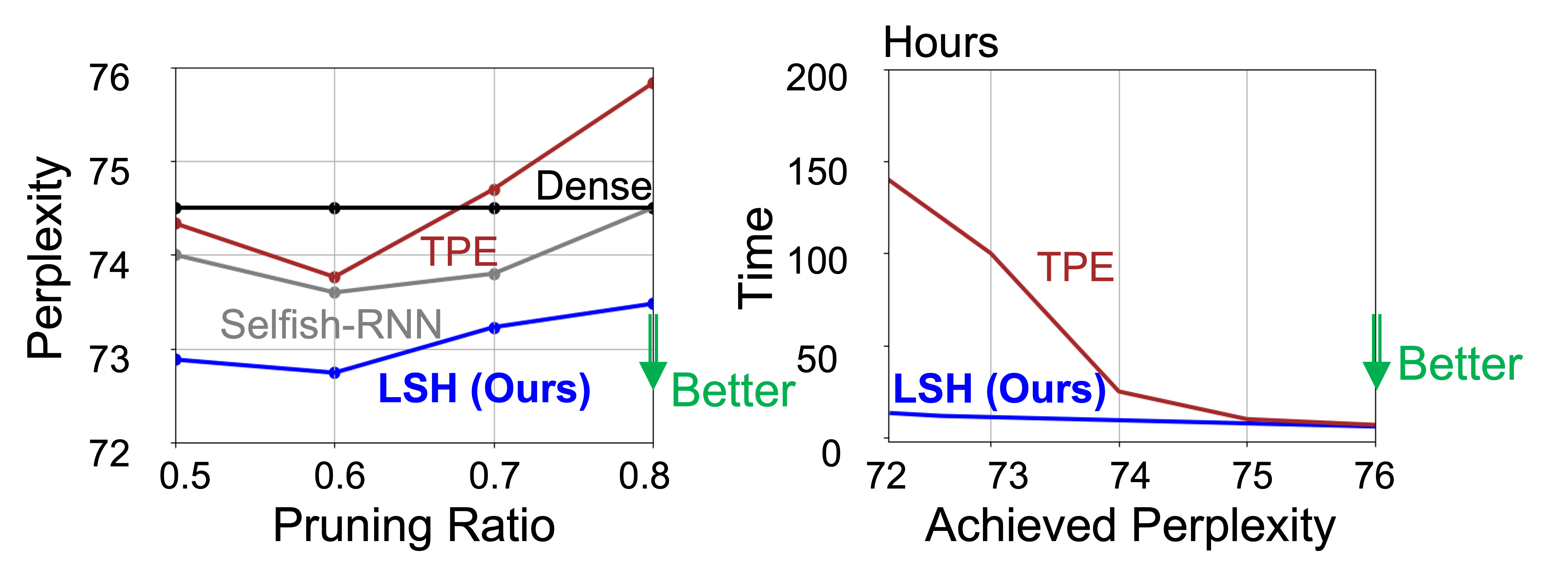}
    \caption{(Left) Pruning Ratio vs. Perplexity: Comparison among LSH (blue), TPE/loss-based (brown), and selfish-RNN (gray) on different pruning ratios.; (Right) Achieved Perplexity vs Time Budget: Comparison between LSH (blue) and loss-based full training search (brown) on required time budget (hours) to reach a target perplexity.}
    \label{fig:result}
\end{figure*}
\subsection*{Robustness}
To evaluate robustness, we examine the performance of LSH under varying pruning ratios, ranging from moderate ($0.5$) to high ($0.8$) with a step size $0.1$. Each experiment uses the stacked LSTM model with a fixed removal/insertion schedule ($E = 3$) and candidate set size of $20$. TPE is used as the underlying optimization algorithm.

As illustrated in Fig.~\ref{fig:result}-Left, LSH (in blue) consistently outperformed TPE/loss-based (in brown) and Selfish-RNN (in gray) methods across all pruning ratios. LSH also consistently achieved higher accuracy than the dense networks across all pruning ratios, showcasing its robustness.

\subsection*{Time Efficiency}
To estimate the computational efficiency of LSH, we compared the time required for LSH to reach a specific target perplexity with the baseline loss-based full training search method, where candidates are fully trained sequentially until one candidate reaches the target perplexity. We used stacked LSTM with a fixed pruning ratio of $0.67$ for these experiments. Fig.~\ref{fig:result}-Right shows the time (y-axis) required to find a configuration with a particular target perplexity (x-axis).

Our observations indicated that LSH outperforms loss-based approach in terms of computational efficiency across all perplexities. Specifically, for finding the optimal configuration (validation perplexity of 72), LSH took approximately $15$ hours, whereas the loss-based approach required approximately $150$ hours ( $> 6$ days). This $10\times$ improvement in efficiency demonstrates LSH’s advantage in rapidly identifying high-quality configurations.

\begin{table}[ht]
\centering
\caption{Comparison between LS-based (LSH) and Loss-based distances applied with GS, TPE, ATPE, Hyperband, and BOHB under different budgets. Stacked LSTM with a pruning ratio of $0.67$ trained on PTB dataset is used here. The average perplexity with $95\%$ confidence interval are reported.\label{tab:LSTM_budgets}}
\begin{tabular}{l|c|c|c} 
\toprule
Methods & Low Budget & Moderate~~Budget & High~~Budget \\
\midrule
GS & 72.5 & 72.5 & 72.5 \\
BOHB & 72.4 $\pm$ 0.5 & 72.5 $\pm$ 0.2 & 72.7 $\pm$ 0.4 \\
TPE & 72.3 $\pm$0.8 & 72.4 $\pm$ 1.0 & 72.7 $\pm$ 1.1 \\
Hyperband & 72.2 $\pm$0.4 & 72.4 $\pm$ 0.2 & 72.4 $\pm$ 0.4 \\
ATPE & 72.0 $\pm$1.9 & 72.7 $\pm$ 0.1 & 72.7 $\pm$ 0.1 \\
LSH-Hyperband & 71.6 $\pm$1.6 & 70.3 $\pm$ 2.0 & 70.7 $\pm$ 0.8 \\
LSH-BOHB & 71.3 $\pm$0.9 & 71.5 $\pm$ 1.2 & 70.1 $\pm$ 0.5 \\
LSH-TPE & 70.9 $\pm$3.0 & 70.9 $\pm$ 2.6 & 69.9 $\pm$ 2.6 \\
LSH-ATPE & 70.9 $\pm$ 0.4 & \textbf{69.9 $\pm$ 0.3} & \textbf{69.9 $\pm$ 0.3} \\
LSH-GS & \textbf{70.8} & 70.8 & 70.8 \\
\bottomrule
\end{tabular}
\end{table}

\subsection*{LS-based vs. loss-based.}


To better understand the computational trade-offs of the proposed method, we conduct ablation studies focusing on two factors: (1) the computational cost of Lyapunov Spectrum (LS) estimation, and (2) the scheduling hyperparameter $E$ that controls the frequency of candidate removal and generation in LSH.

We covered various hyperparameter optimization algorithms, including Grid Search (GS), TPE, Adaptive TPE (APTE), Hyperband, and BOHB. For these experiments, we employed the same stacked LSTM with a pruning ratio of $0.67$ as in previous experiments. In the case of GS, we fixed the death rate to $0.8$ and only considered methodological hyperparameters, resulting in $24$ configurations. For other hyperparameter optimization methods, we tested LS-based and loss-based distances under different resource budgets, each defined by a different initial candidate set size, namely low, moderate, and high resource scenarios, with $24$, $30$, and $40$ candidates in the initial set, respectively. Each experiment was repeated thrice, and the average testing perplexity with a $95\%$ confidence interval was reported.

As demonstrated in Table.~\ref{tab:LSTM_budgets}, LS-based distance (bottom 5 rows) consistently reached lower perplexity compared to loss-based distance counterparts (top 5 rows) across all resource budgets. This demonstrates that LS-based distance appears to be a more effective early estimation criterion than loss at that time of training. Notably, while LS-distance coupled with advanced hyperparameter optimization algorithms such as ATPE and BOHB did not match the performance of GS counterpart when resources were limited (1st column), they outperformed the GS counterpart when more resources were provided (3rd column). For example, with an increased resource budget (Low → High), LSH improved the average performance, with ATPE improving from $70.9$ to $69.9$ and BOHB from $71.3$ to $70.1$. Additionally, LSH achieved a more reliable result as the confidence interval decreased from $0.4$ to $0.3$ for ATPE and from $0.9$ to $0.5$ for BOHB. In contrast, this improvement was not consistently observed in the loss-based counterpart, with mean perplexities remaining relatively unchanged ($72.3$ to $72.7$ for TPE and $72.2$ to $72.4$ for Hyperband). These results indicate that LS-based distance leverages advanced hyperparameter optimization methods more effectively, resulting in more optimal elimination and generation processes and ultimately leading to better selections.

\subsection*{Ablation Study}
In this section, we conducted ablation studies to assess the computational cost of LS computation and the effect of scheduling for the completeness of the LSH approach. These experiments follow the same setup as the previous stacked LSTM experiments.

LS computation proved computationally expensive and memory-intensive. Due to the need to compute QR decomposition and Jacobian matrix multiplication, this additional computational effort, beyond network training, could slow the selection when many samples were used.  In our experiments, each sample required approximately 6 seconds LS computation. 

\begin{table}[ht]
\centering
\caption{Effect of LS computation batch size: The maximum, mean, and variance of LS computed from $2$ and $10$ validation samples on two pruned models.}
\label{tab:num_samples_vs_LE}
\begin{tabular}{l|ccc|ccc}
\toprule
\multirow{2}{*}{Stats}    & \multicolumn{3}{c|}{Pruned 1} & \multicolumn{3}{c}{Pruned 2}  \\
         & 2     & 10    & Delta        & 2     & 10    & Delta         \\
\midrule
Max      & -1.26 & -1.20 & 0.06         & -0.52 & -0.46 & 0.06          \\
Min      & -4.40 & -4.41 & 0.01         & -3.97 & -3.97 & 0             \\
Variance & 2.59  & 2.67  & 0.08         & 1.92  & 1.92  & 0             \\
\bottomrule
\end{tabular}
\end{table}

To address this computational cost, we conducted experiments to empirically evaluate whether accurate LS computation could be achieved using only a few validation samples. In particular, we compared the maximum, mean, and variance of LS computed from different numbers of validation samples across two pruned networks. These features were identified as critical characteristics of LS~\cite{10.3389/fams.2022.818799}. Table.~\ref{tab:num_samples_vs_LE} shows that the maximum, mean, and variance of LS computed from $2$ and $10$ validation samples were similar. The difference ($\Delta$) between these cases was an order of magnitude smaller than the difference observed between pruned networks. Based on this observation, we use only $2$ samples for LS computation, reducing the computational time to approximately $12$ seconds.

Our findings revealed that while LS computation was significant, it only added a $20\%$ overhead to the candidate selection process compared to the loss-based method when considering the overall time spent on training a single epoch with SGD ($60$ seconds) as the baseline. Furthermore, it is important to note that LS computation did not impact the extensive training process significantly. In a previous experiment with 40 initial candidates, the overall time increased by only $7.5\%$ ($861$ minutes compared to $801$ minutes). This highlights that while LS computation could add computational cost, only a few samples were required, making the overall computational overhead reasonable, especially for larger problems. Additionally, for larger datasets such as WikiText-2, where training AWD-LSTM-MoS for one epoch took over 600 seconds (much longer than computing LS, which took 5 seconds), the benefits of LSH are more pronounced.

\begin{table}[ht]
    \centering
    \caption{The efficiency and effectiveness of scheduling, including time for selection procedures, extensive training, and the entire process.}
    \label{tab:schedule_effect}
    \begin{tabular}[c]{l|c|c|c|c}
    \toprule
    $E$ & 1 & 2 & 3 & 4 \\
    \midrule
    selection & 45 & 90 & 135 & 180 \\
    \midrule
    extensive training & 590 & 580 & 570 & 560 \\ 
    \midrule
    total time (s) & 635 & 670 & 705 & 740\\
    \midrule
    optimal config selected & No & No & \textbf{Yes} & \textbf{Yes}\\
    \bottomrule
    \end{tabular}
    \label{table:scheduling}
\end{table}

The hyperparameter $E$, defining the searching schedule, also significantly impacts the efficiency and effectiveness of LSH. For this study, we utilized Grid Search as the search method, considering three aforementioned methodological hyperparameters and resulting in 24 different configurations. We tuned the incremental epochs parameter ($E$) to explore various schedules. Since the number of available hyperparameter configurations is limited, only candidate removal is used, and candidate generation is not applied. Table.~\ref{table:scheduling} shows the time required for the selection procedures, extensive training, and the entire process. Note that only when $E \ge 3$ the optimal configuration is successfully selected for extensive training. While it increases the entire time by 35 minutes as $E$ increases from $2$ to $3$, this is relatively small compared to the entire time ($\approx$ 5\%).

\section{Conclusion}
In this study, we introduced a novel hyperparameter search to select the optimal configuration, including the pruning method and its corresponding hyperparameters for specific scenarios. In particular, we proposed the Lyapunov Spectrum (LS)-based Hyperpruning algorithm, referred to as LSH, which is designed to effectively and efficiently identify the optimal configurations. LSH removes less promising pruned variants based on their distance to a dense reference network in the LS space and subsequently generates enhanced variants in an iterative way. More resources are then allocated to the more promising remaining candidates, allowing for the extensive training of optimally pruned networks. To evaluate the performance of LSH, we conducted experiments on stacked LSTM and RHN models trained with the Penn Treebank Dataset and on AWD-LSTM-MoS with the WikiText-2 Dataset for language modeling tasks. Our experimental results demonstrate that LSH consistently performs better than other state-of-the-art pruning methods and even dense networks. Moreover, the performance of LSH is robust across different pruning ratios, and it achieves an order of magnitude boost in terms of search time compared to a full training search. LSH method can be seamlessly integrated into various existing hyperparameter optimization algorithms, such as TPE and ATPE, and it is superior to traditional loss-based distance as a metric for those hyperparameter optimization algorithms. In summary, LSH offers a promising avenue for hyperparameter optimization in the context of neural network pruning, offering significant efficiency improvements and ultimately leading to the selection of optimal pruned networks.

\newpage
\begin{appendices}

\include{supplementary}

\end{appendices}

\bibliography{ref}

\end{document}

%% file: supplementary.tex
\section{Hyperparameter Values}
\label{sm:hyperparameter}
We show all hyperparameters related to Stacked LSTM, RHN, and AWD-LSTM-MoS experiments in Table.~\ref{tab:hyperparameter_list}. Hyperpruning does not optimize these hyperparameters.

\begin{table*}[b]
\centering
\caption{Hyperparameters list: dataset (Data), dimension of hidden units of each layer (H-dim), number of layers (Layers), dimension of input embedding (Emb), Optimization (Opt), Learning rate (LR), Non-monotone interval for SNT-ASGD (Non-mono), Training batch size (BS), Back-Propagation Through Time (BPTT), Dropout for word-level, embedding, hidden layer and output (Dropout), Training epochs (Epochs), Gradient Clip (Clip), LS computation batch size (LS-BS), Sparsity decay schedule (Decay Sche), Encoder and Decoder weight tied (Tied), Carry gate and the transform gate coupled (Coupled).}
\label{tab:hyperparameter_list}
\begin{tabular}{l|c|c|c}
\toprule
\textbf{Model}    & S\textbf{tacked-LSTM} & \textbf{RHN}   & \textbf{AWD-LSTM-MoS}       \\
\midrule
Data     & PTB          & PTB   & WikiText-2         \\
H-Dim    & (1500, 1500) & (830) & (1150, 1150, 650)  \\
Layers   & 2            & 1     & 3                  \\
Emb      & 1500            & 830                     & 830                    \\
Opt      & SNT-ASGD        & SNT-ASGD                & SNT-ASGD               \\
LR       & 40              & 15                      & 15                     \\
Non-Mono & 5 & 5 & 5 \\
BS       & 20 & 20 & 15 \\
BPTT     & 35 & 35 & 70 \\
Dropout  & (0, 0, 0.65, 0) & (0.2, 0.65, 0.25, 0.62) & (0.1, 0.55, 0.2, 0.4)  \\
Epochs   & 100             & 500                     & 1000                   \\
Clip     & 0.25 & 0.25 & 0.25 \\
LS-BS    & 2 & 2 & 2 \\
Decay    & Cosine & Cosine & Cosine \\
Tied     & False           & True                    & False                  \\
Coupled  & False           & True                    & False                  \\
\bottomrule
\end{tabular}
\end{table*}

\section{Embedding Spaces}
\label{sm:embed}
Our work employed PCA as the embedding function and $L_2$ distance as the distance metric. However, it is important to note that there are no special constraints on this particular embedding function and distance metric. 
To thoroughly investigate the completeness of LSH, we conducted ablation studies on various embedding spaces and distance measurements in the embedding space. The ablation study aimed to understand LSH and its performance in different scenarios comprehensively.

We use the following equations to calculate the distance between the candidate network $c^i$ in the candidate pool $\mathcal{P}$ and the reference dense network $\hat{c}$: 
\begin{align*}
\centering
    [\hat{v}, v^i_0, \dots, v^i_{-1}] & = embedding([\hat{\Lambda}, \Lambda^i])\\
    s & = distance(\hat{v}, v^i_{-1}),\\
\end{align*}
where $\Lambda^i_j$ stands for the LS of the $i$-th candidate at the $j$-th epoch and $\Lambda^i$ groups all LS of $c^i$ till current epoch, i.e.,$\Lambda^i \equiv \{\Lambda^i_j\}_{j=0}$. $\hat{\Lambda}$ denotes the LS of the reference network. The embedded LS of reference $\hat{c}$ and candidate $c^i$ at the current epoch, i.e., $\hat{v}$ and $v^i_{-1}$, is used to calculate the distance.

We conducted an ablation study using three embedding spaces: PCA, original LS, and T-SNE, along with two distance metrics: $L_2$ and Cosine distance, which is defined below.
\begin{align*}
    L_2(p, q) & = \sqrt{\sum_{i=1}^n (p_i - q_i)^2}\\
    Cos(p, q) & = 1 - \frac{p \cdot q}{|p||q|} = 1 - \frac{\sum_{i=1}^n (p_i q_i)}{\sqrt{\sum_{i=1}^n (p_i)^2} + \sqrt{\sum_{i=1}^n (q_i)^2}},\\
\end{align*}
Table.~\ref{tab:emd_dist} presents the testing perplexities obtained by the selected network based on these various combinations of embedding functions and distance metrics. Each combination performs reasonably, demonstrating the effectiveness of using different embedding spaces and distance metrics within the LSH framework. Surprisingly, a similar performance was achieved even when using the original LS with the Cosine distance metric. This finding suggests that the LS is informative regarding the network's performance even at the early stages of training. Among the different combinations, we observed that PCA with L2 distance, the configuration used in LSH, is preferable over other combinations. This led to the choice of the PCA embedding function combined with the L2 distance metric in our experiments.

\begin{table}[ht]
    \centering
    \caption{Perplexities of the final model selected based on different combinations of embedding spaces and distance metrics.}
    \label{tab:emd_dist}
    \begin{tabular}[c]{|c|cc|cc|cc|}
        \hline
        Embedding  & \multicolumn{2}{c|}{PCA}                   & \multicolumn{2}{c|}{LS}           & \multicolumn{2}{c|}{T-SNE}       \\ \hline
        Distance   & \multicolumn{1}{c|}{L2}            & Cos   & \multicolumn{1}{c|}{L2}    & Cos  & \multicolumn{1}{c|}{L2}   & Cos  \\ \hline
        Perplexity & \multicolumn{1}{c|}{\textbf{69.9}} & 72.99 & \multicolumn{1}{c|}{72.56} & 70.3 & \multicolumn{1}{c|}{70.5} & 72.6 \\ \hline
    \end{tabular}
\end{table}

\section{Jacobian Matrix Derivation}
\label{sm:jacobian}
\subsection{Stacked LSTM}
The LSTM is defined by the following equations:
\begin{align*}
    f_t & = \sigma_g(W_f x_t + U_f h_{t-1} + b_f)\\
    i_t & = \sigma_g(W_i x_t + U_i h_{t-1} + b_i)\\
    o_t & = \sigma_g(W_o x_t + U_o h_{t-1} + b_o)\\
    c^{\prime}_t & = \sigma_c(W_c x_t + U_c h_{t-1} + b_c)\\
    c_t & = f_t \cdot c_{t-1} + i_t \cdot c^{\prime}_t\\
    h_t & = o_t \cdot \sigma_c(c_t),\\
\end{align*}

where $\sigma_g$ and $\sigma_c$ are element-wise sigmoid function and $tanh$ function, respectively. For simplicity, we use the following notation when calculating the derivatives:

\begin{align*}
    y_* & = W_* x_t + U_* h_{t-1} + b_{*},\\
\end{align*}
where $*$ can be ${f, i, o, c}$, depending on the gate. The derivative of each of those gates/states with respect to the hidden states is shown below:

\begin{align*}
    \frac{\partial f_t}{\partial h_{t-1}} & = [\sigma_g(y_f) (1 - \sigma_g(y_f))]^T U_f\\
    \frac{\partial i_t}{\partial h_{t-1}} & = [\sigma_g(y_i) (1 - \sigma_g(y_i))]^T U_i\\
    \frac{\partial o_t}{\partial h_{t-1}} & = [\sigma_g(y_o) (1 - \sigma_g(y_o))]^T U_o\\
    \frac{\partial c_t}{\partial h_{t-1}} & = \frac{\partial f_t}{\partial h_{t-1}} c_{t-1} + \frac{\partial i_t}{\partial h_{t-1}} tanh(y_c) + i_t sech^2(y_c) U_c\\
    \frac{\partial h_{t}}{\partial h_{t-1}} & = \frac{\partial o_t}{\partial h_{t-1}} tanh(c_t) + o_t sect^2(c_t) \frac{c_t}{\partial h_{t-1}},\\
\end{align*}

The derivative of each of those gates/states with respect to the inputs is shown below:

\begin{align*}
    \frac{\partial f_t}{\partial x_t} & = [\sigma_g(y_f) (1 - \sigma_g(y_f))]^T W_f\\
    \frac{\partial i_t}{\partial x_t} & = [\sigma_g(y_i) (1 - \sigma_g(y_i))]^T W_i\\
    \frac{\partial o_t}{\partial x_t} & = [\sigma_g(y_o) (1 - \sigma_g(y_o))]^T W_o\\
    \frac{\partial c_t}{\partial x_t} & = \frac{\partial f_t}{\partial x_t} c_{t-1} + \frac{\partial i_t}{\partial x_t} tanh(y_c) + i_t sech^2(y_c) W_c\\
    \frac{\partial h_{t}}{\partial x_t} & = \frac{\partial o_t}{\partial x_t} tanh(c_t) + o_t sect^2(c_t) \frac{\partial c_t}{\partial x_t},\\
\end{align*}

\subsection{RHN}
An RHN layer with a recurrent depth of $L$ is described by:

\begin{align*}
    h^{[t]}_{l} & = p^{[t]}_{l} \cdot e^{[t]}_{l} + h^{[t]}_{l-1} \cdot c^{[t]}_{l}, \\
    & l \in [1, L],
\end{align*}
where

\begin{align}
    p^{[t]}_{l} & = tanh(W_P x^{[t]}\mathbbm{1}_{\{l=1\}} + R_{P_l} h_{l-1} ^{[t]} + b_{P_l})\\
    e^{[t]}_{l} & = \sigma(W_E x^{[t]}\mathbbm{1}_{\{l=1\}} + R_{E_l} h_{l-1} ^{[t]} + b_{E_l})\\
    c^{[t]}_{l} & = \sigma(W_C x^{[t]}\mathbbm{1}_{\{l=1\}} + R_{C_l} h_{l-1} ^{[t]} + b_{C_l})
\end{align}

and $\mathbbm{1}$ is the indicator function. The output of the RNH layer is the output of the $L^{th}$ Highway layer, i.e., $y^{[t]} = h_L^{[t]}$. The derivative is shown below:

\begin{align*}
    \frac{h_1^{[t]}}{h_1^{[t-1]}} & = \frac{\partial h_1^{[t]}}{\partial h_L^{[t-1]}} \prod_{l=1}^{L-1} \frac{\partial h_{l+1}^{[t-1]}}{\partial h_{l}^{[t-1]}}\\
\end{align*}

where
\begin{align*}
    \frac{\partial h_{l+1}^{[t-1]}}{\partial h_{l}^{[t-1]}} & = P_{l+1}^{\prime[t-1]} diag(e_{l+1}^{[t-1]}) + E_{l+1}^{\prime[t-1]} diag(p_{l+1}^{[t-1]}) + diag(c_{l+1}^{[t-1]}) + C_{l+1}^{\prime[t-1]} diag(h_l^{[t-1]}),
\end{align*}
with
\begin{align*}
    P_{l+1}^{\prime[t-1]} & = R_{P_{l+1}}^T diag[tanh^{\prime}(R_{P_{l+1}} h_{l} ^{[t-1]} + b_{P_{l+1}})]\\
    E_{l+1}^{\prime[t-1]} & = R_{E_{l+1}}^T diag[tanh^{\prime}(R_{E_{l+1}} h_{l} ^{[t-1]} + b_{E_{l+1}})]\\
    C_{l+1}^{\prime[t-1]} & = R_{C_{l+1}}^T diag[tanh^{\prime}(R_{C_{l+1}} h_{l} ^{[t-1]} + b_{C_{l+1}})],\\
\end{align*}

and
\begin{align*}
    \frac{\partial h_1^{[t]}}{\partial h_L^{[t-1]}} & = P_1^{\prime[t]} diag(e_1^{[t]}) + E_1^{\prime[t]} diag(p_1^{[t]}) + diag(c_1^{[t]}) + C_1^{\prime[t]} diag(h_L^{[t-1]}) \\
\end{align*}
with
\begin{align*}
    P_1^{\prime[t]} & = R_{P_1}^T diag[tanh^{\prime}(W_P x^{[t]} + R_{P_1} h_{L} ^{[t-1]} + b_{P_1})]\\
    E_1^{\prime[t]} & = R_{E_1}^T diag[tanh^{\prime}(W_E x^{[t]} + R_{E_1} h_{L} ^{[t-1]} + b_{E_1})]\\
    C_1^{\prime[t]} & = R_{C_1}^T diag[tanh^{\prime}(W_C x^{[t]} + R_{C_1} h_{L} ^{[t-1]} + b_{C_1})]\\
\end{align*}

If it is coupled, c = 1 - e. The RHN becomes 

\begin{align*}
    h^{[t]}_{l}. = p^{[t]}_{l} \cdot e^{[t]}_{l} + h^{[t]}_{l-1} \cdot (1 - e^{[t]}_{l})
\end{align*}